\documentclass[twoside,11pt]{article}

\usepackage{jmlr2e}

\usepackage{times}
\usepackage{color}
\usepackage{amsmath}
\usepackage{bm}
\usepackage{subfigure}
\usepackage{url}
\usepackage{multirow}
\usepackage{subfigure}
\usepackage{threeparttable}

\jmlrheading{X}{20XX}{XX-XX}{XX/XX}{XX/XX}{meila00a}{Hongteng Xu and Hongyuan Zha}

\ShortHeadings{THAP: A Matlab Toolkit for Hawkes Processes}{Xu and Zha}
\firstpageno{1}

\begin{document}

\title{THAP: A Matlab Toolkit for Learning with Hawkes Processes}

\author{\name Hongteng Xu \email hxu42@gatech.edu \\
       \addr School of Electrical and Computer Engineering\\
       Georgia Institute of Technology\\
       Atlanta, GA 30332, USA
       \AND
       \name Hongyuan Zha \email zha@cc.gatech.edu \\
       \addr College of Computing\\
       Georgia Institute of Technology\\
       Atlanta, GA 30332, USA}

\editor{XX}

\maketitle

\begin{abstract}
As a powerful tool of asynchronous event sequence analysis, point processes have been studied for a long time and achieved numerous successes in different fields. 
Among various point process models, Hawkes process and its variants attract many researchers in statistics and computer science these years because they capture the self- and mutually-triggering patterns between different events in complicated sequences explicitly and quantitatively and are broadly applicable to many practical problems.
In this paper, we describe an open-source toolkit implementing many learning algorithms and analysis tools for Hawkes process model and its variants. 
Our toolkit systematically summarizes recent state-of-the-art algorithms as well as most classic algorithms of Hawkes processes, which is beneficial for both academical education and research. 
Source code can be downloaded from \url{https://github.com/HongtengXu/Hawkes-Process-Toolkit}.
\end{abstract}


\section{Introduction}
Real-world interactions among multiple entities are often recorded as event sequences, such as user behaviors in social networks, earthquakes in different locations, and diseases and their complications. 
The entities or event types in these sequences often exhibit complicated self- and mutually-triggering patterns --- historical events are likely to have influences on the happenings of current and future events, and the historical events at different time stamps have different impacts. 
Modeling these event sequences and analyzing the triggering patterns behind them are classical problems in statistics and computer science, which can be solved based on point process models and their learning algorithms. 

As a special kind of point processes, Hawkes process model~\citep{hawkes1971point} attracts a lot of researchers and has been widely used in many fields because it can represent the triggering patterns explicitly and quantitatively. 
Additionally, Hawkes process is very flexible, which has many variants and can be extended and connected with existing machine learning models. 
Its typical applications include, but not limited to, financial analysis, bioinformatics, social network analysis and control, and crowd behavior modeling.
Because of these properties and broad applications, most existing toolkits of point processes are actually developed focusing on Hawkes processes.

However, although many new models and learning algorithms of Hawkes processes have been proposed for these years, the development of existing Hawkes processes' toolkits lags behind. 
On the one hand, they concentrate on implementing traditional algorithms rather than the rapidly evolving state-of-the-art. 
On the other hand, it is difficult to have a fair and comprehensive comparison for modern algorithms because they are implemented over different sources.

Focusing on modeling and learning Hawkes process and its variants, we describe a new toolkit \textit{THAP} (\textbf{T}oolkit for \textbf{HA}wkes \textbf{P}rocesses) in this paper, implementing a wide variety of learning and analysis algorithms for Hawkes processes. 
THAP offers a Matlab-based implementation of modern state-of-the-art learning and analysis algorithms and provides two real-world date sets (the IPTV data~\citep{luo2014you,luo2015multi} and the Linkedin data~\citep{xu2017learning}). 
It has an ability to compare different algorithms using a variety of evaluation metrics, and thus, may clarify which algorithms perform better under what circumstance. 
The Matlab-based implementation makes it have some benefits for academical education and research --- students can understand the basic concepts of Hawkes processes and the details of the corresponding learning algorithms and accelerate their research in their initial phases. 
The open-source nature of \textit{THAP} makes it easy for third parties to contribute additional implementations, and the modules of \textit{THAP} are extendable to develop more complicated models and functions, e.g. Wasserstein learning~\citep{xiao2017wasserstein} and recurrent neural networks~\citep{du2016recurrent}. 

\begin{figure}[!t]
\centering
\includegraphics[height=4.1cm,width=1\textwidth]{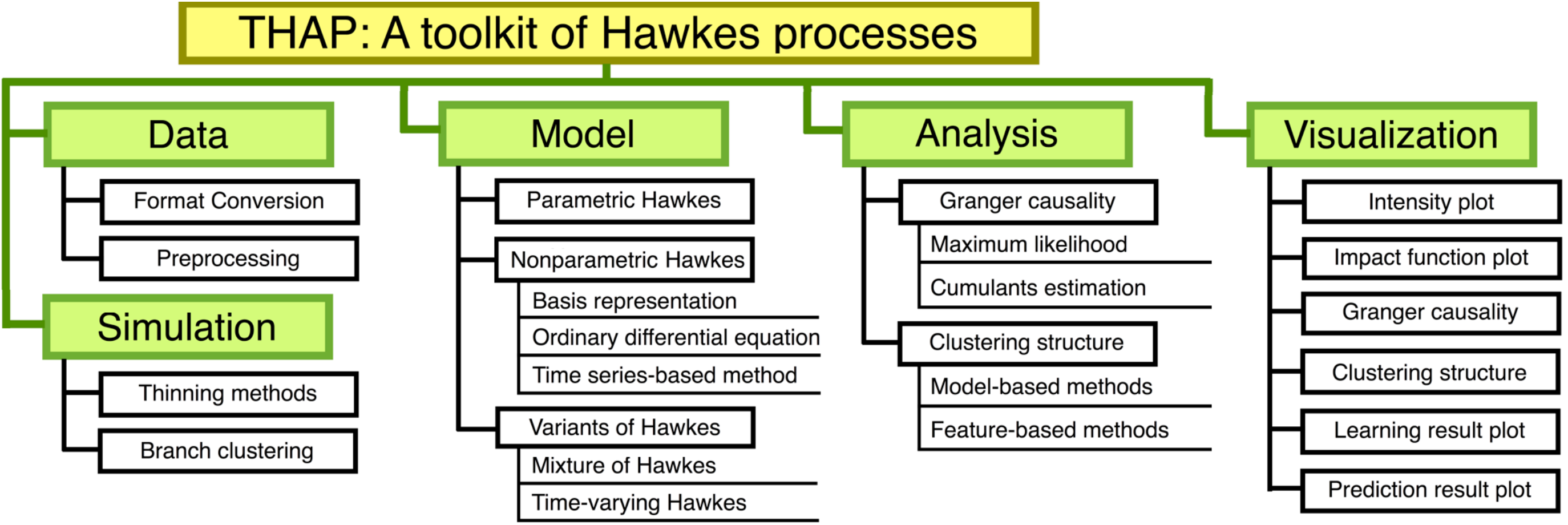}\vspace{-7pt}
\caption{The structure of THAP.}\label{Fig1}
\end{figure}

\section{Implementation}
\textit{THAP} is a multi-platform Matlab software (R2016a or higher version required). 
It is compatible with MS Windows, Linux, and Mac OS. 
The toolkit consists of five main components, as shown in Fig.~\ref{Fig1}.
\textbf{Data:} Import real-world data (i.e., csv files), convert them to Matlab's format (i.e., mat files), and implement data preprocessing like sampling, stitching, and thinning.
\textbf{Simulation:} Implement three simulation methods to generate synthetic data, including the branch clustering method~\citep{hawkes1974cluster,moller2006approximate}, Ogata's modified thinning method~\citep{ogata1981lewis}, and the fast thinning method for the Hawkes process with exponential impact functions~\citep{dassios2013exact}.
\textbf{Model:} Define Hawkes process model and its variants and implement their learning algorithms.
\textbf{Analysis:} Achieve the Granger causality analysis and the clustering analysis of event sequences.
\textbf{Visualization:} Visualize data, models, and learning results.

The key modules of \textit{THAP} are modeling and analysis modules. 
In particular, Hawkes processes can be categorized into parametric models and nonparametric ones. 
The parametric models include the Hawkes processes with predefined impact functions, e.g., exponential impact functions and Gaussian impact functions. 
The nonparametric models include the Hawkes processes with arbitrary impact functions, and those impact functions can be represented by a set of basis functions or discretized as a set of sample points with fixed time lags. 
According to the representation of impact function, different learning algorithms are applied. 
For the Hawkes processes with continuous impact functions (i.e., those represented by predefined functions or basis), we can apply maximum likelihood estimation (MLE) directly to estimate the parameters of the models.
For the Hawkes processes with discretized impact functions, we can (a) treat event sequences as time series and apply the least-squares (LS) method~\citep{eichler2016graphical}, or (b) combine MLE with a solver of ordinary differential equations (ODE)~\citep{zhou2013learning2} to estimate the parameters of the models. 

\begin{figure}[t]
\centering
\subfigure[Sequence and intensity]{
\includegraphics[height=2.6cm,width=0.22\linewidth]{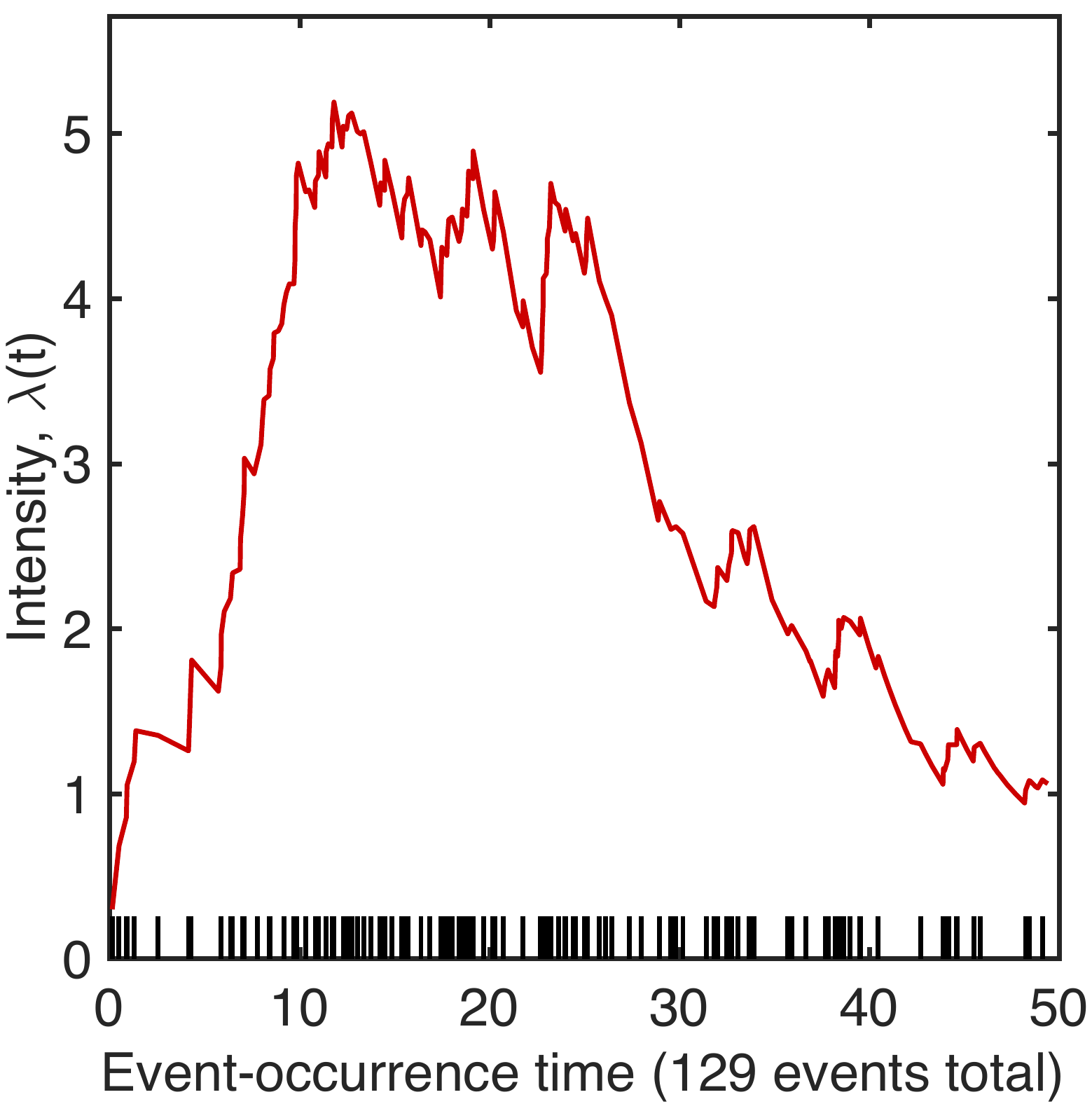}
}
\subfigure[Simulators' runtime]{
\includegraphics[height=2.6cm,width=0.22\linewidth]{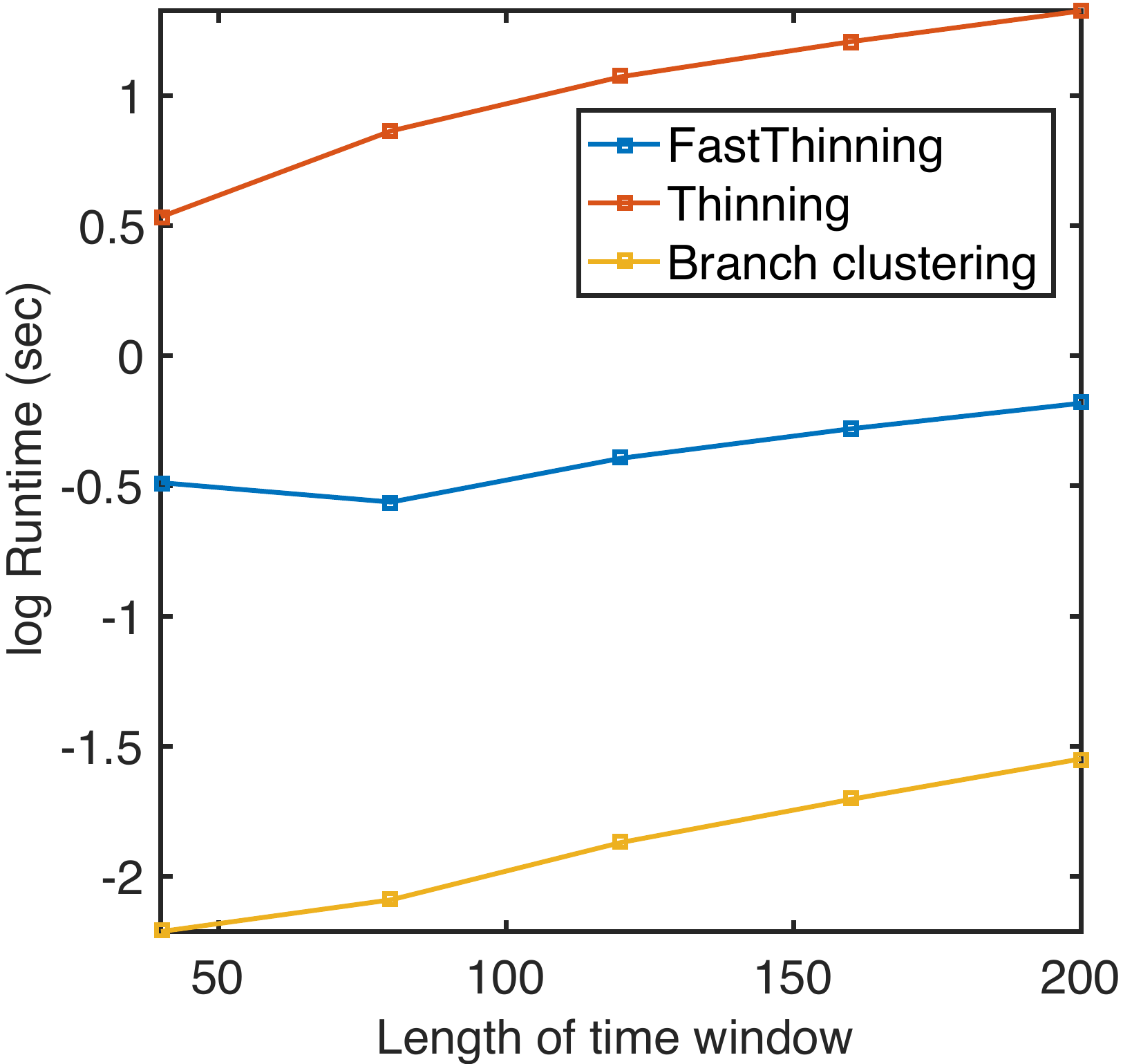}
}
\subfigure[Learned impact function]{
\includegraphics[height=2.6cm,width=0.22\linewidth]{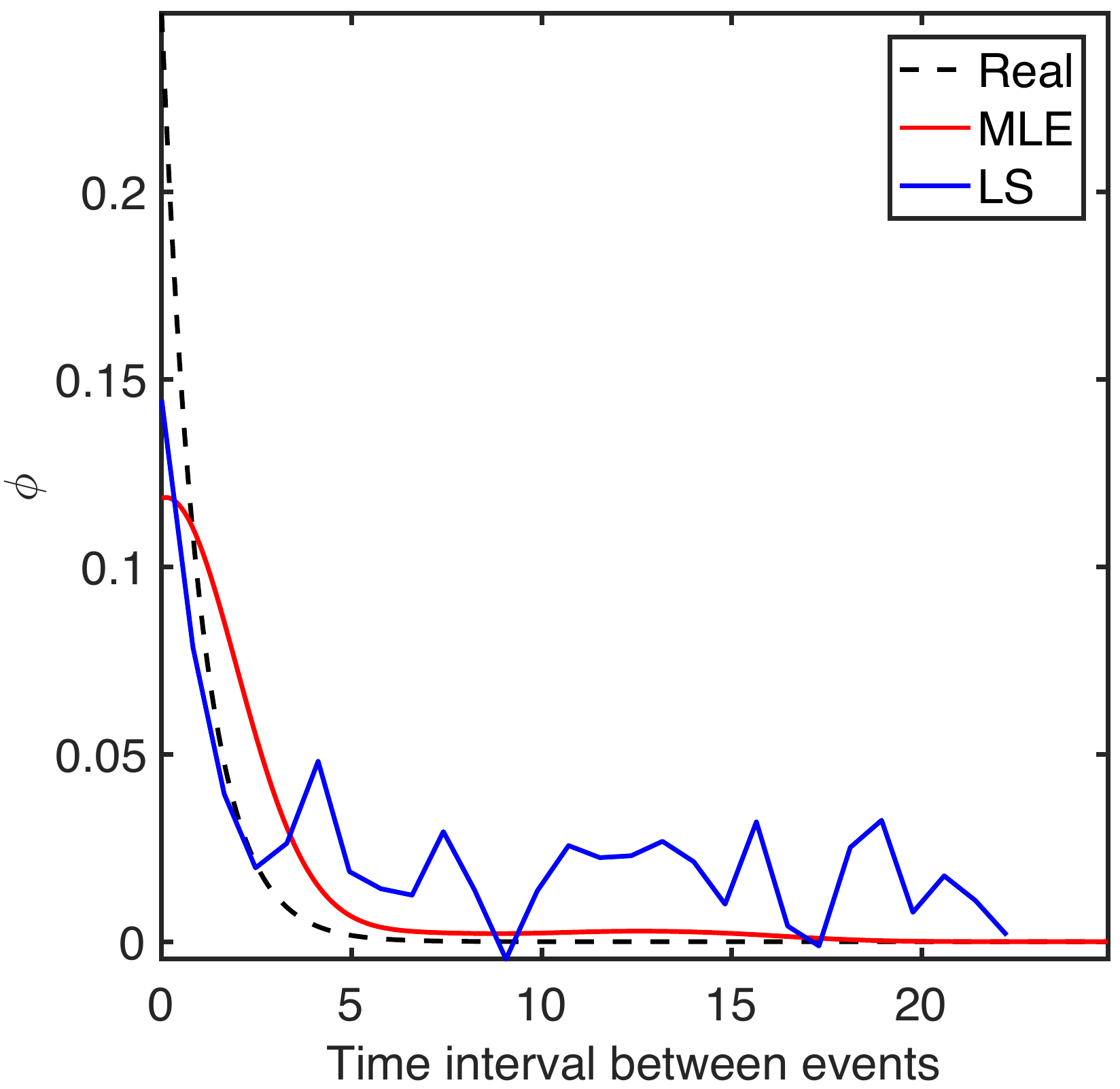}
}
\subfigure[Estimation errors]{
\includegraphics[height=2.6cm,width=0.22\linewidth]{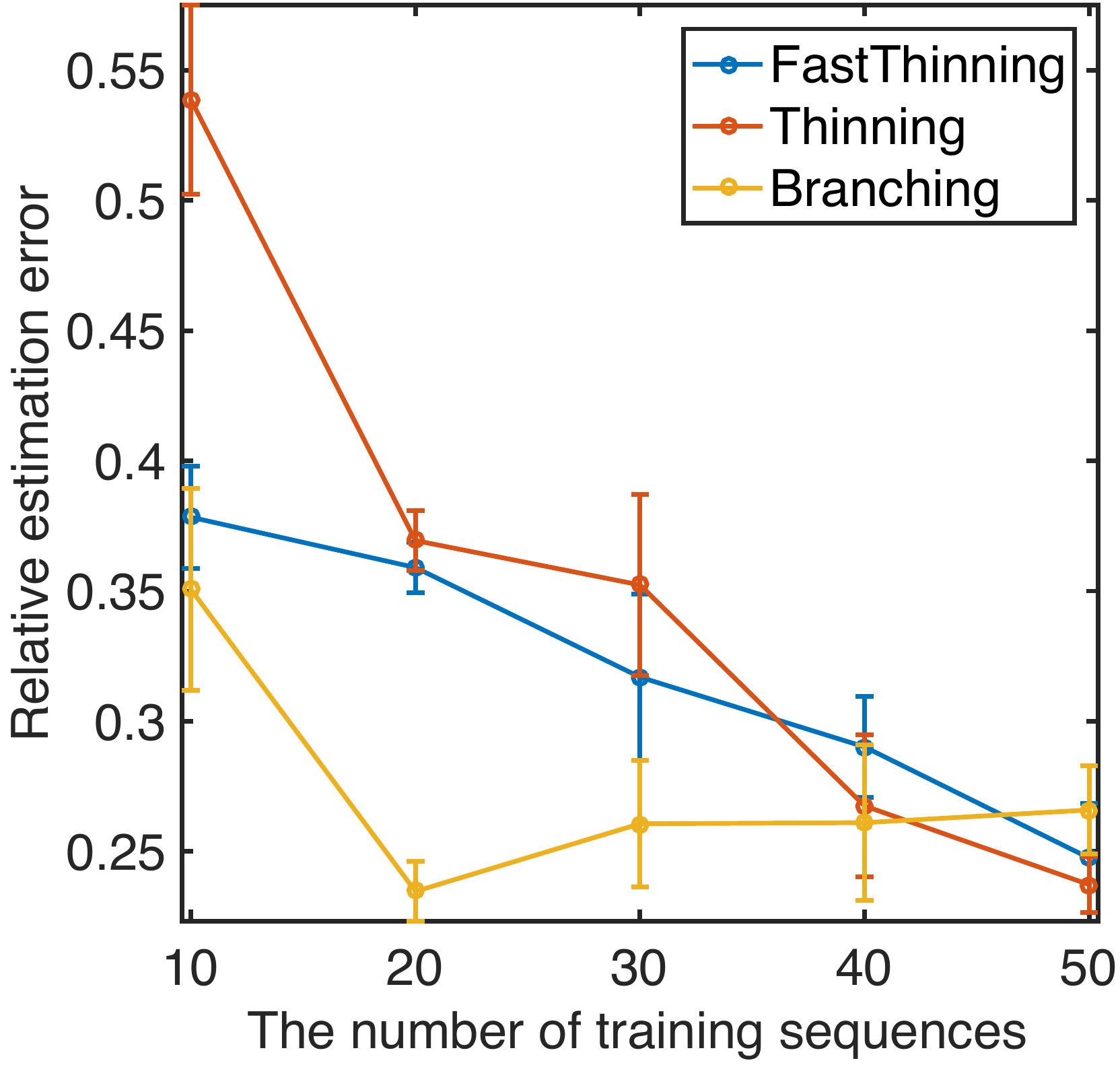}
}\\
\subfigure[Log-likelihood of data]{
\includegraphics[height=2.6cm,width=0.22\linewidth]{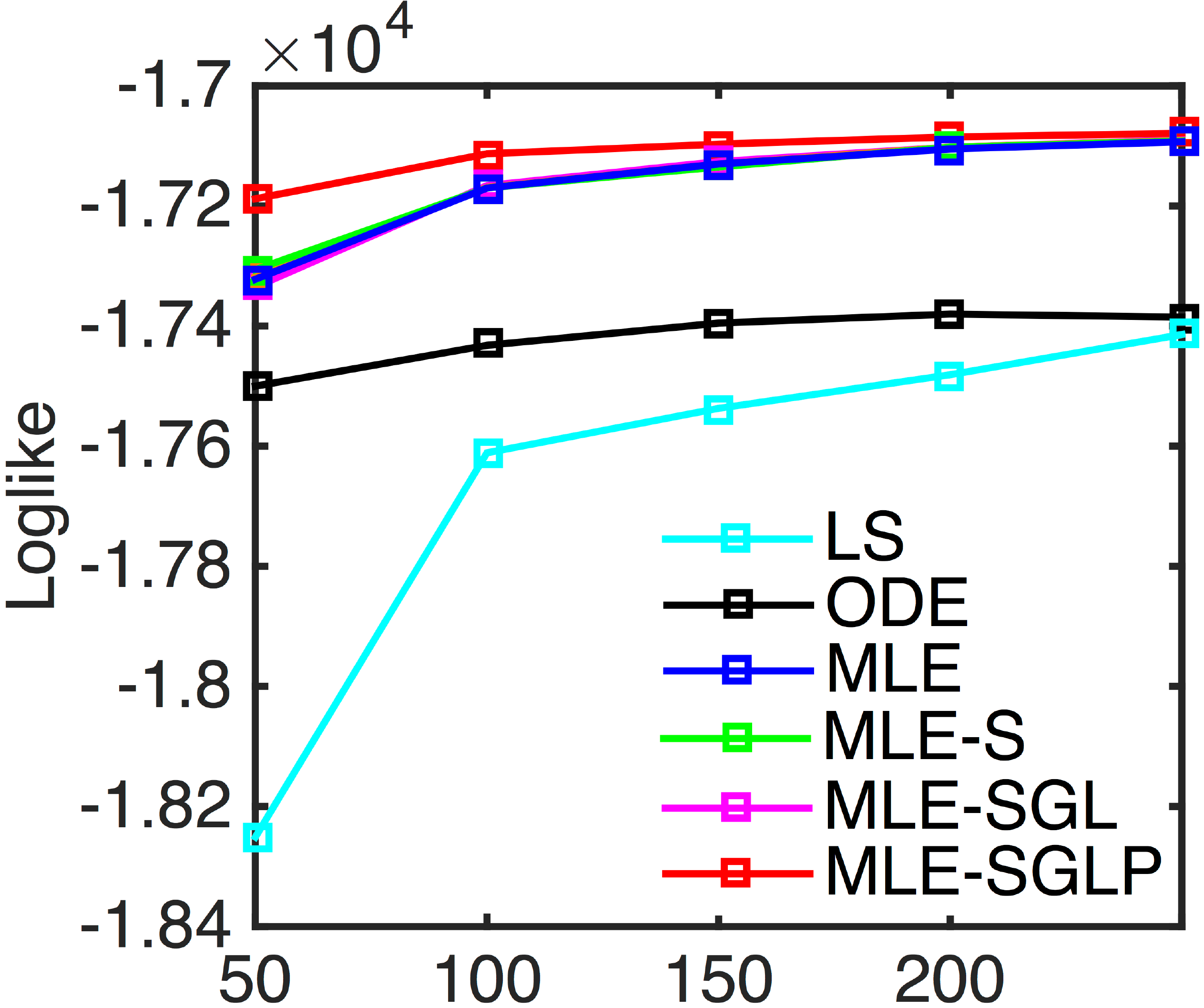}
}
\subfigure[Granger causality graph]{
\includegraphics[height=2.6cm,width=0.22\linewidth]{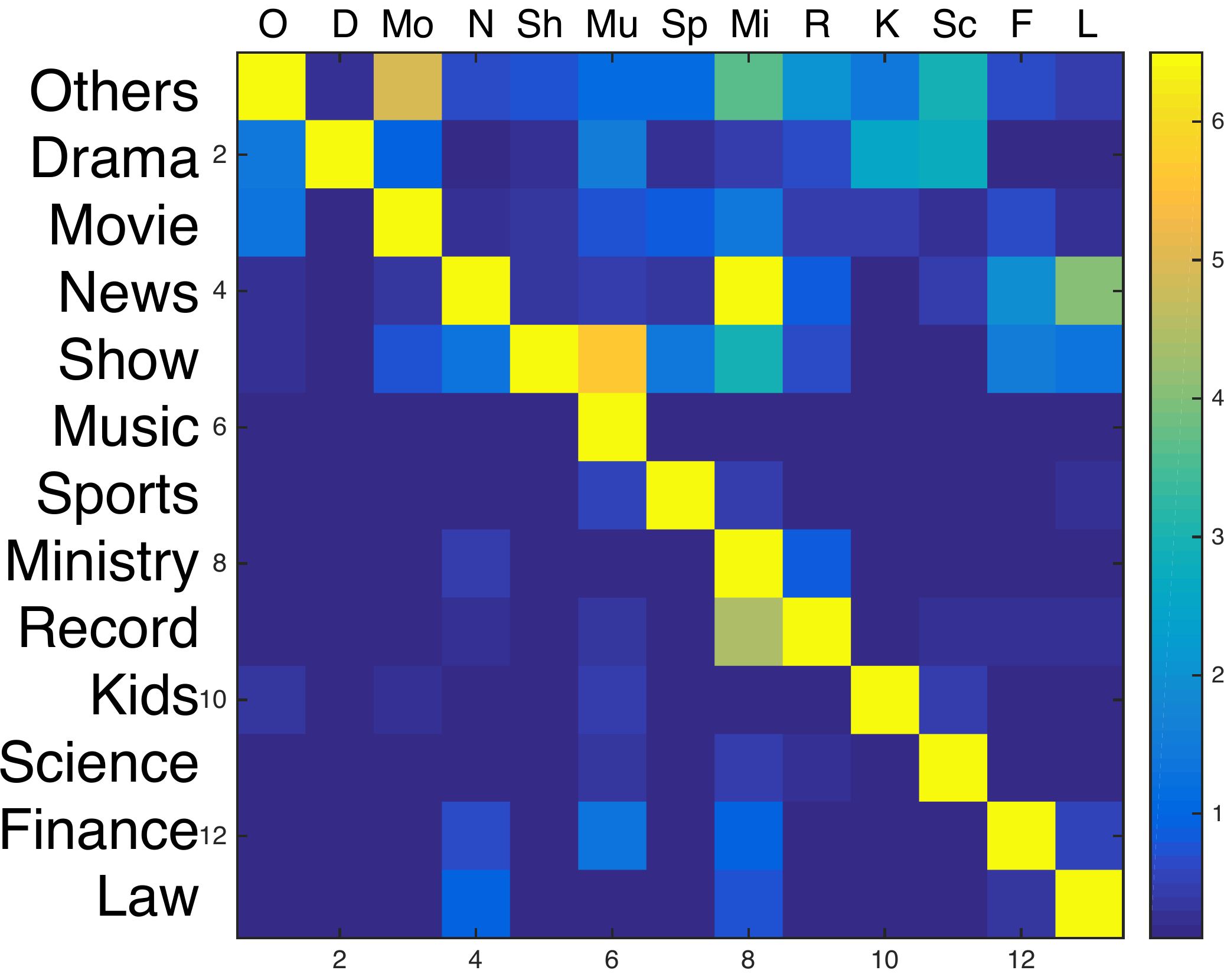}
}
\subfigure[Dynamics of infectivity]{
\includegraphics[height=2.6cm,width=0.22\linewidth]{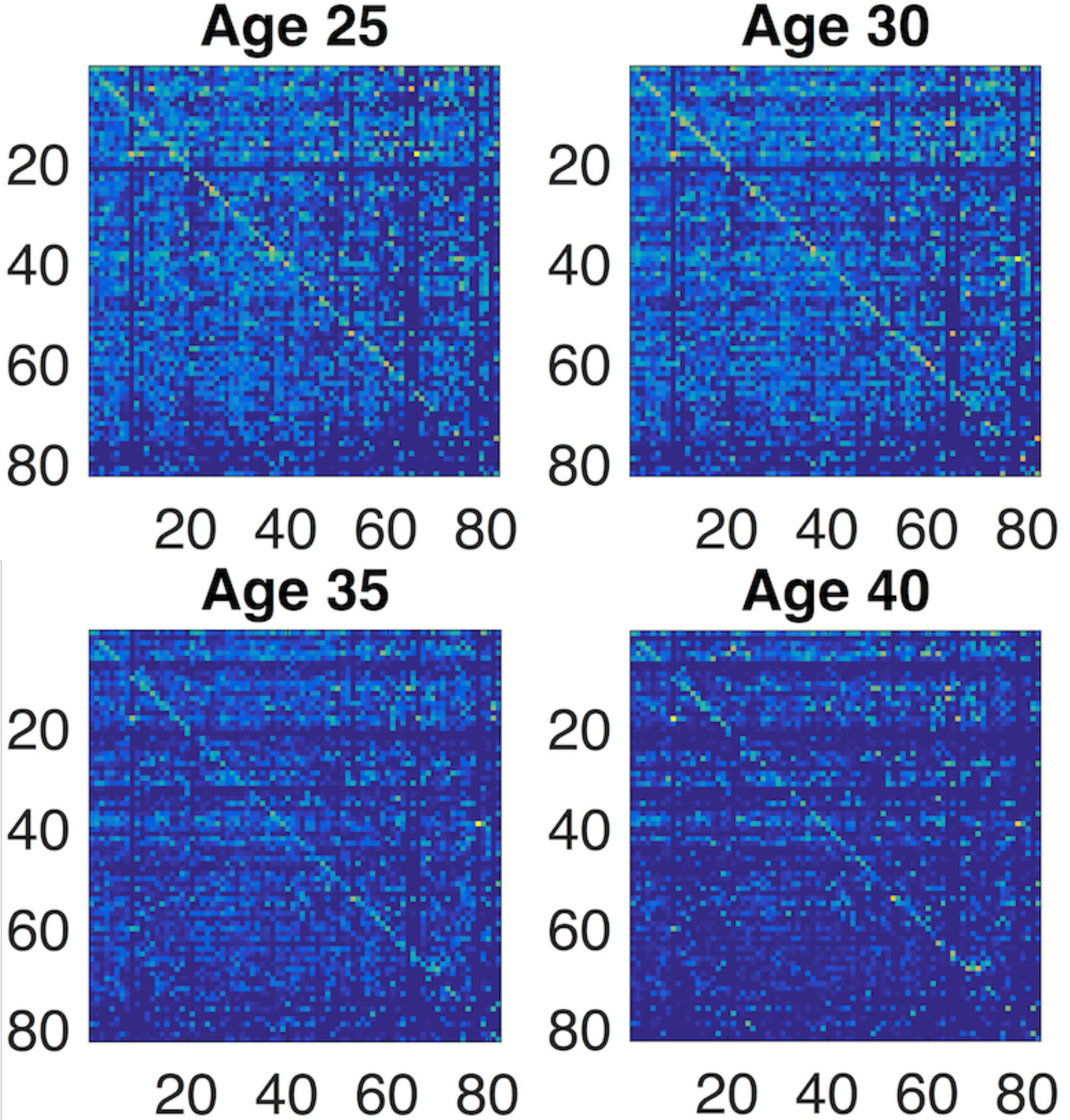}
}
\subfigure[Clustering analysis]{
\includegraphics[height=2.6cm,width=0.22\linewidth]{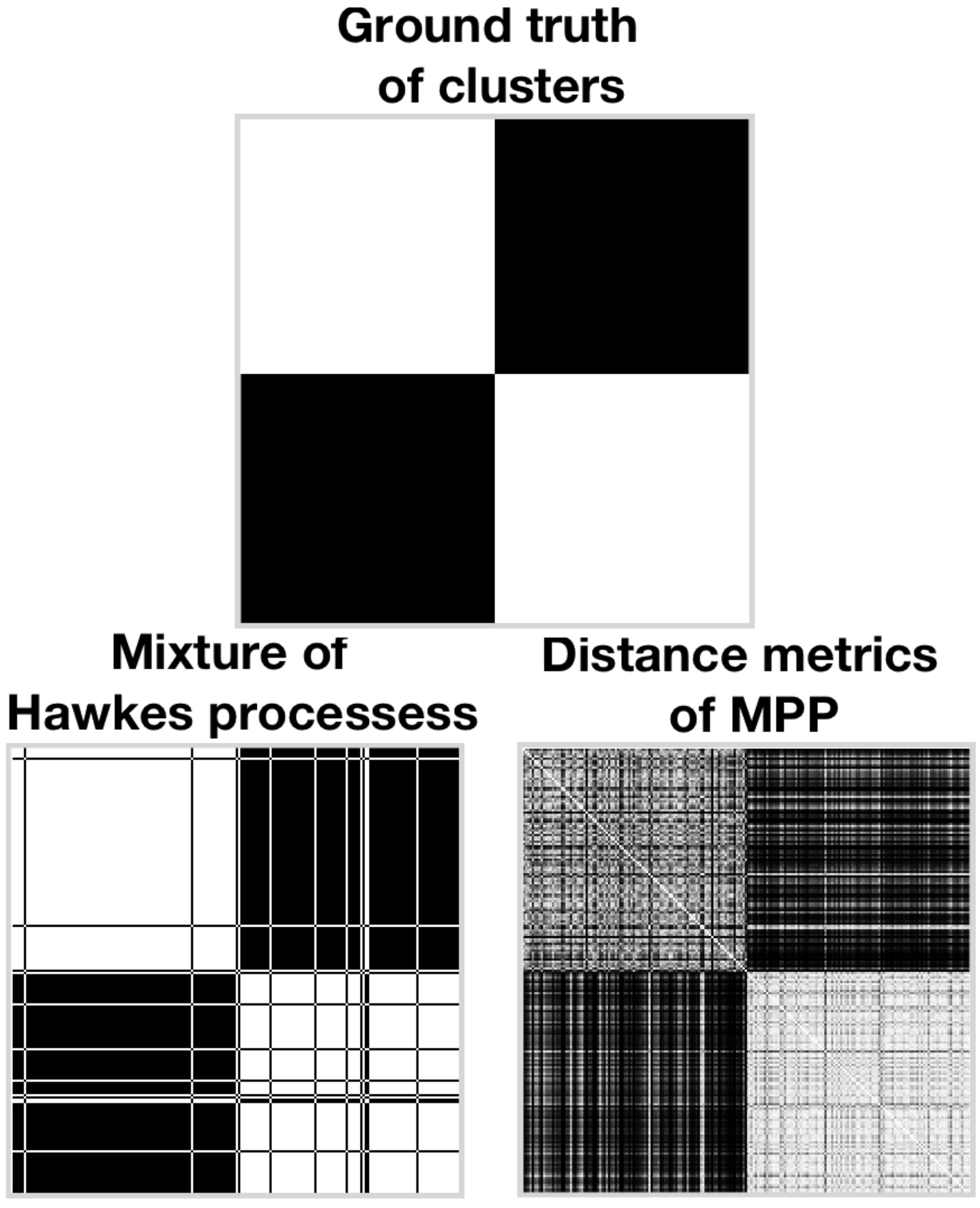}
}\vspace{-7pt}
\caption{Visualization of typical functions achieved by \textit{THAP}}\label{Fig2}
\end{figure}

Additionally, \textit{THAP} provides us with three cutting-edge analysis tools. 
The first is \textbf{Granger causality analysis.} 
For multi-dimensional Hawkes processes, the self-and mutually-triggering patterns between different event types can be represented by a Granger causality graph. 
\textit{THAP} combines MLE with various regularizers, e.g., sparse, group-sparse, and low-rank regularizers, and learns the adjacent matrix of the Granger causality graph (i.e., the infectivity matrix of event types) robustly. 
The second is \textbf{clustering analysis.} 
\textit{THAP} contains two methods to cluster the event sequences generated by different Hawkes processes: (a) learning a mixture model of Hawkes processes~\citep{xu2017dirichlet}; 
(b) implementing a distance metric for marked point processes~\citep{iwayama2017definition}.
The third is \textbf{dynamical analysis of event sequence.} 
\textit{THAP} implements the time-varying Hawkes process (TVHP) model in~\citep{xu2017learning} and captures the change of infectivity matrix over time.

In summary, we visualize some typical functions achieved by \textit{THAP} in Fig.~\ref{Fig2}. 
Specifically, using \textit{THAP}, we can (a) visualize event sequences and their intensity functions; (b) simulate event sequences by different simulators and compare their runtime; (c) learn Hawkes processes by different algorithms and visualize learned impact functions; (d) calculate estimation errors of parameters; (e) calculate log-likelihood of data obtained by different algorithms; (f) learn the Granger causality graph of event types (e.g., the infectivity between TV program categories); (g) learn the dynamics of infectivity matrix (e.g., the infectivity between companies for employees at different ages); and (h) learn clustering structures of event sequences and distances between them.

\section{Related work}
Table~\ref{tab1} summarizes the implemented features in other open-source point process toolkits and compares them to those in \textit{THAP}. 
The functions implemented by different toolkits are labeled by different symbols. 
\textit{THAP} covers most of functions of other toolkits and contains many new functions. 
Specifically, the R-based library \textit{R-hawkes}\footnote{\url{ https://cran.r-project.org/web/packages/hawkes/hawkes.pdf}} just contains a single estimation algorithm of traditional Hawkes processes~\citep{da2014hawkes}. 
The Python-based library \textit{pyhawkes}\footnote{\url{https://github.com/slinderman/pyhawkes}} only implements its contributors' published algorithms~\citep{linderman2014discovering,linderman2015scalable}. 
The C++ library \textit{PtPack}\footnote{\url{ https://github.com/dunan/MultiVariatePointProcess}}
includes some traditional and advanced learning techniques of point processes. 
It is not very user-friendly because it does not have Python or Matlab interfaces. 
Recently, a new C++ library \textit{tick}\footnote{\url{https://github.com/X-DataInitiative/tick}} is developed with a Python interface~\citep{bacry2017tick}, which includes most \textit{PtPack}'s functions and further improves their performance.

\begin{table}[!t]
  \centering
  \small
  \caption{Models and algorithms of Hawkes processes in different toolkits.\label{tab1}}
  \begin{threeparttable}[c]
      \begin{tabular}{
        c|c|c|c|c|c 
        }
        \hline\hline
        \multirow{2}{*}{Model} & Type & \multicolumn{2}{c|}{Parametric} &\multicolumn{2}{c}{Nonparametric}\\ \cline{2-6}
        & Impact function &Exponential  &~~Gaussian\quad 	&Smooth basis &~~Discrete\quad\\
        \hline
        \multirow{2}{*}{Simulator}& Branch clustering &$\bigstar\blacksquare\clubsuit$ &$\bigstar\clubsuit$ &$\bigstar\clubsuit$ &\\
        & (Fast) Thinning &$\bigstar\blacklozenge\clubsuit\spadesuit$ &$\bigstar\clubsuit\spadesuit$ &$\bigstar\clubsuit\spadesuit$ &\\ \hline
        \multirow{3}{*}{Learning} & MLE(+Regularizer) & $\bigstar\blacksquare\clubsuit\spadesuit$ &$\bigstar\clubsuit\spadesuit$ &$\bigstar\clubsuit\spadesuit$ &\\
        &MLE + ODE &$\bigstar\clubsuit\spadesuit$ &$\bigstar\clubsuit\spadesuit$ &$\bigstar\clubsuit\spadesuit$ &$\bigstar\clubsuit\spadesuit$\\
        &Least-squares &$\bigstar$ & & &$\bigstar$\\ 
        \hline
        \multirow{4}{*}{Analysis} & Granger causality & $\bigstar\blacksquare\clubsuit\spadesuit$ &$\bigstar\clubsuit\spadesuit$ &$\bigstar\clubsuit\spadesuit$ &$\spadesuit$\\
        & Clustering (Mixture model) & $\bigstar$ &$\bigstar$ &$\bigstar$ &\\
        & Clustering (Distance metric) &$\bigstar$ &$\bigstar$ &$\bigstar$ &$\bigstar$\\
        & Longtime dynamics (TVHP) & $\bigstar\clubsuit$ &$\bigstar\clubsuit$ &$\bigstar\clubsuit$ &\\
        \hline\hline
      \end{tabular}
  \end{threeparttable}
  \begin{tablenotes}
  		\begin{scriptsize}
         \item $\bigstar=$ \textit{THAP}, $\blacklozenge=$ \textit{R-hawkes}, $\blacksquare=$ \textit{pyhawkes}, $\clubsuit=$ \textit{PtPack}, $\spadesuit=$ \textit{tick}.\par
  		\end{scriptsize}
  \end{tablenotes}
\end{table}

\section{Summary}
\textit{THAP} contributes to point process research community by (a) providing
an easy and fair comparison among most existing models and learning algorithms of Hawkes processes, (b) supporting
advanced analysis tools which have not been available for other libraries, and (c) filling the blank of point process's education and research with a Matlab-based toolkit.
In the future, we plan to add extensions to go beyond existing Hawkes process models. 

\acks{We would like to acknowledge support for this project
from the NSF IIS-1639792, 1717916, and NSFC 61628203.}


\newpage

%
%
%

\vskip 0.2in
\begin{small}
\bibliography{sample}

\begin{thebibliography}{18}
\providecommand{\natexlab}[1]{#1}
\providecommand{\url}[1]{\texttt{#1}}
\expandafter\ifx\csname urlstyle\endcsname\relax
  \providecommand{\doi}[1]{doi: #1}\else
  \providecommand{\doi}{doi: \begingroup \urlstyle{rm}\Url}\fi

\bibitem[Bacry et~al.(2017)Bacry, Bompaire, Ga{\"\i}ffas, and
  Poulsen]{bacry2017tick}
Emmanuel Bacry, Martin Bompaire, St{\'e}phane Ga{\"\i}ffas, and Soren Poulsen.
\newblock tick: a python library for statistical learning, with a particular
  emphasis on time-dependent modeling.
\newblock \emph{arXiv preprint arXiv:1707.03003}, 2017.

\bibitem[Da~Fonseca and Zaatour(2014)]{da2014hawkes}
Jos{\'e} Da~Fonseca and Riadh Zaatour.
\newblock Hawkes process: Fast calibration, application to trade clustering,
  and diffusive limit.
\newblock \emph{Journal of Futures Markets}, 34\penalty0 (6):\penalty0
  548--579, 2014.

\bibitem[Dassios and Zhao(2013)]{dassios2013exact}
Angelos Dassios and Hongbiao Zhao.
\newblock Exact simulation of hawkes process with exponentially decaying
  intensity.
\newblock \emph{Electronic Communications in Probability}, 18\penalty0
  (62):\penalty0 1--13, 2013.

\bibitem[Du et~al.(2016)Du, Dai, Trivedi, Upadhyay, Gomez-Rodriguez, and
  Song]{du2016recurrent}
Nan Du, Hanjun Dai, Rakshit Trivedi, Utkarsh Upadhyay, Manuel Gomez-Rodriguez,
  and Le~Song.
\newblock Recurrent marked temporal point processes: Embedding event history to
  vector.
\newblock In \emph{KDD}, pages 1555--1564. ACM, 2016.

\bibitem[Eichler et~al.(2016)Eichler, Dahlhaus, and
  Dueck]{eichler2016graphical}
Michael Eichler, Rainer Dahlhaus, and Johannes Dueck.
\newblock Graphical modeling for multivariate hawkes processes with
  nonparametric link functions.
\newblock \emph{Journal of Time Series Analysis}, 2016.

\bibitem[Hawkes(1971)]{hawkes1971point}
Alan Hawkes.
\newblock Point spectra of some mutually exciting point processes.
\newblock \emph{Journal of the Royal Statistical Society. Series B
  (Methodological)}, pages 438--443, 1971.

\bibitem[Hawkes and Oakes(1974)]{hawkes1974cluster}
Alan Hawkes and David Oakes.
\newblock A cluster process representation of a self-exciting process.
\newblock \emph{Journal of Applied Probability}, pages 493--503, 1974.

\bibitem[Iwayama et~al.(2017)Iwayama, Hirata, and
  Aihara]{iwayama2017definition}
Koji Iwayama, Yoshito Hirata, and Kazuyuki Aihara.
\newblock Definition of distance for nonlinear time series analysis of marked
  point process data.
\newblock \emph{Physics Letters A}, 381\penalty0 (4):\penalty0 257--262, 2017.

\bibitem[Linderman and Adams(2014)]{linderman2014discovering}
Scott Linderman and Ryan Adams.
\newblock Discovering latent network structure in point process data.
\newblock In \emph{ICML}, pages 1413--1421, 2014.

\bibitem[Linderman and Adams(2015)]{linderman2015scalable}
Scott Linderman and Ryan Adams.
\newblock Scalable bayesian inference for excitatory point process networks.
\newblock \emph{arXiv preprint arXiv:1507.03228}, 2015.

\bibitem[Luo et~al.(2014)Luo, Xu, Zha, Du, Xie, Yang, and Zhang]{luo2014you}
Dixin Luo, Hongteng Xu, Hongyuan Zha, Jun Du, Rong Xie, Xiaokang Yang, and
  Wenjun Zhang.
\newblock You are what you watch and when you watch: Inferring household
  structures from iptv viewing data.
\newblock \emph{IEEE Transactions on Broadcasting}, 60\penalty0 (1):\penalty0
  61--72, 2014.

\bibitem[Luo et~al.(2015)Luo, Xu, Zhen, Ning, Zha, Yang, and
  Zhang]{luo2015multi}
Dixin Luo, Hongteng Xu, Yi~Zhen, Xia Ning, Hongyuan Zha, Xiaokang Yang, and
  Wenjun Zhang.
\newblock Multi-task multi-dimensional hawkes processes for modeling event
  sequences.
\newblock In \emph{IJCAI}, 2015.

\bibitem[M{\o}ller and Rasmussen(2006)]{moller2006approximate}
Jesper M{\o}ller and Jakob Rasmussen.
\newblock Approximate simulation of hawkes processes.
\newblock \emph{Methodology and Computing in Applied Probability}, 8\penalty0
  (1):\penalty0 53--64, 2006.

\bibitem[Ogata(1981)]{ogata1981lewis}
Yosihiko Ogata.
\newblock On lewis' simulation method for point processes.
\newblock \emph{IEEE Transactions on Information Theory}, 27\penalty0
  (1):\penalty0 23--31, 1981.

\bibitem[Xiao et~al.(2017)Xiao, Farajtabar, Ye, Yan, Song, and
  Zha]{xiao2017wasserstein}
Shuai Xiao, Mehrdad Farajtabar, Xiaojing Ye, Junchi Yan, Le~Song, and Hongyuan
  Zha.
\newblock Wasserstein learning of deep generative point process models.
\newblock \emph{arXiv preprint arXiv:1705.08051}, 2017.

\bibitem[Xu and Zha(2017)]{xu2017dirichlet}
Hongteng Xu and Hongyuan Zha.
\newblock A dirichlet mixture model of hawkes processes for event sequence
  clustering.
\newblock \emph{arXiv preprint arXiv:1701.09177}, 2017.

\bibitem[Xu et~al.(2017)Xu, Luo, and Zha]{xu2017learning}
Hongteng Xu, Dixin Luo, and Hongyuan Zha.
\newblock Learning hawkes processes from short doubly-censored event sequences.
\newblock In \emph{ICML}, 2017.

\bibitem[Zhou et~al.(2013)Zhou, Zha, and Song]{zhou2013learning2}
Ke~Zhou, Hongyuan Zha, and Le~Song.
\newblock Learning triggering kernels for multi-dimensional hawkes processes.
\newblock In \emph{ICML}, 2013.

\end{thebibliography}
\end{small}

\end{document}